\documentclass[journal]{IEEEtran}
\usepackage{graphicx} 

\usepackage{multirow}
\usepackage{algorithm}
\usepackage{algpseudocode}
\usepackage{amsmath}
\usepackage{amssymb}
\usepackage{amsfonts}
\usepackage{hyperref}
\usepackage{mathrsfs} 
\usepackage{bm}
\usepackage{booktabs} 
\usepackage{xcolor} 
\usepackage{subfig}

\usepackage{verbatim} 

\newcommand{\fault}{D}
\newcommand{\rate}{\dot{\fault}}
\newcommand{\trans}[1]{#1'}
\newcommand{\tuple}[1]{( #1 )}
\newcommand{\obsscalar}{x}
\newcommand{\obs}{\mathbf{\obsscalar}}
\newcommand{\ts}{\mathbf{X}}
\newcommand{\tsset}{\mathbb{X}}
\newcommand{\real}{\mathbb{R}}
\newcommand{\nsens}{M}
\newcommand{\nsamp}{N}
\newcommand{\nts}{T}
\newcommand{\out}{y}
\newcommand{\outset}{\bm{\out}}

\begin{document}
\title{A textual transform of multivariate time-series for prognostics}
\author{Abhay~Harpale,
  Abhishek~Srivastav
  \thanks{A. Harpale and A. Srivastav are with the Machine Learning Lab, GE Global Research, San Ramon, CA. Email: \texttt{\{harpale, srivastav\}@ge.com}}}

\maketitle

\begin{abstract}\small\baselineskip=9pt
  Prognostics or early detection of incipient faults is an important industrial
  challenge for condition-based and preventive maintenance. Physics-based
approaches to modeling fault progression are infeasible due to multiple
interacting components, uncontrolled environmental factors and 
observability constraints. Moreover,
such approaches to prognostics do not generalize to new domains. Consequently, domain-agnostic data-driven machine learning
approaches to prognostics are desirable.
Damage progression is a path-dependent process and
explicitly modeling the temporal patterns is critical for accurate estimation of
both the current damage state and its progression leading to total failure. In this paper, we present a novel data-driven approach to
prognostics that employs a novel textual representation of multivariate temporal
sensor observations for predicting the future health state of the monitored
equipment early in its life. This representation enables us to utilize
well-understood concepts from text-mining for modeling, prediction and
understanding distress patterns in a domain agnostic way. The approach has been deployed and successfully tested on large scale multivariate time-series data from commercial aircraft engines. We
report experiments on
well-known publicly available benchmark datasets and simulation datasets. The proposed approach is shown to be superior in terms of
prediction accuracy, lead time to prediction and interpretability. 
\end{abstract}

\section{Introduction}\label{sec:intro}

\IEEEPARstart{I}{ndustrial} equipment such as aircraft engines, locomotives and gas turbines
follow a conservative cadence of scheduled maintenance to insure equipment
availability and safe operations.  It is still common to have several
unscheduled maintenance events that lead to unplanned downtime, loss of
productivity and in some situations jeopardize public safety. Condition-based
maintenance (CBM) can reduce maintenance cost while still providing safe operations. The primary goal of CBM is to track and
predict equipment \emph{health} using sensed measurements
and operating conditions. The latter task of predicting the future health state and estimating remaining useful life falls under the umbrella of \emph{Prognostics}. It has been estimated that the savings achieved from optimizing the use of
engineering equipment can lead to up to \$30B of savings in the aviation
sector alone~\cite{IndustrialInternet}.


Observability into the degradation pattern of an equipment and
likelihood estimate of fault at a future time can greatly improve availability,
efficiency and safety of industrial equipments -- costly maintenances
can be held off until needed, and equipment load or usage can be
adjusted based on current and predicted health status. However,
accurate prognostics for complex engineering systems is challenging
due to (1) interdependent components, (2) lack of full physical
understanding of various failure modes and their effects on
performance, and (3) accurate modeling of damage progression under
uncertain operating conditions. While the nominal operating behavior
of an engineering system can be modeled and controlled, modeling the
response of a faulty equipment or component to operating conditions
and control commands is challenging. In this regard, data-driven
methods, particularly domain-agnostics ones, are a good alternative to
create predictive models of system health.

While ubiquitous sensing and increased connectivity has given greater
visibility into the behavioral changes of a machine as it ages, it has
fueled a surge in massive time-series data from industrial and
commercial equipment such as aircraft engines, gas turbines and home
appliances. In this scenario, the task of modeling \emph{hidden}
health state in a data-driven manner becomes even harder as a large
volume of temporal data needs to be ingested both during the learning
and the real-time application phase. Moreover, during the application
phase these models need to run on low footprint embedded processors to
predict equipment health.

We propose a novel approach for early identification of industrial
units with shorter life spans. Our approach works by transforming the
multivariate temporal data into a sequence of tokenized symbols and
utilizes popular off-the-shelf text-classification strategies for
predictive modeling. Working a in a domain agnostic way, this approach
provides a scalable solution for modeling temporal patterns in sensor
data and predicting the future health state of an industrial
equipment. We use real and simulation data-sets for equipment
degradation to demonstrate our methodology.  The system has been
deployed and tested successfully on real multivariate time-series data
from commercial jet engines acquired over 18 months. In the paper, to
maintain confidentiality and reproducibility of our experiments, we
use the popular aircraft engine degradation benchmark datasets
provided by NASA~\cite{CMAPSS} and degradation or phase-transition of
a chaotic-oscillator~\cite{B97}. Experiments show that our approach is
able to capture the temporal dynamics of damage propagation to
accurately predict the future health state of an equipment early in
its life. We show that our approach is robust in the presence of
significant noise, whereas well-known existing approaches, temporal as
well as non-temporal, perform comparably only when the dataset is much
cleaner.

\subsection*{Notation}

\begin{figure}\label{fig: notation}
  \includegraphics[width=0.95\columnwidth]{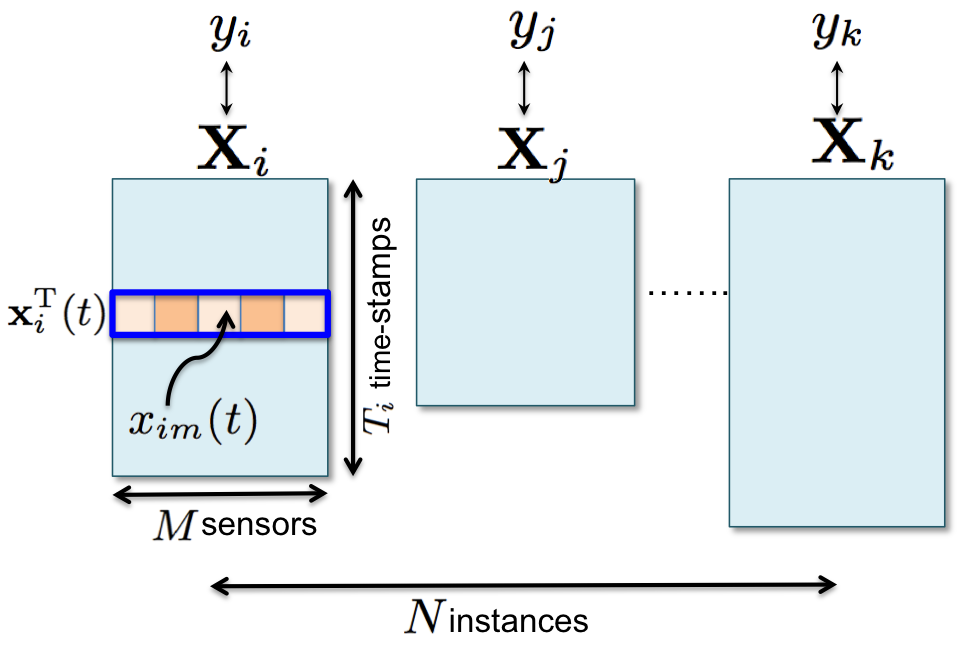}
  \caption{Time-series dataset notation}
  \vspace{-12pt}
\end{figure}

Let $\obs_{i}(t) \in \real^{\nsens}$ be the $\nsens$-dimensional
column vector of observations made at time $t$ for equipment $i$, with
$\obsscalar_{im}(t)$ as the value of the $m^{\text{th}}$ sensor at
time $t$. Let $\ts_{i}=\trans{[\obs_{i}(1), \dots,
    \obs_{i}(\nts_{i})]} = (\obs_i)_0^{\nts_i}$ be a
$\nsens$-dimensional time-series data, where each row corresponds to a
multivariate observation made at a particular time-stamp. Let the
collection $\tsset=\{\ts_{j}: 1\leq j \leq \nsamp\}$ be the
time-series dataset of $\nsamp$ instances. Let $\outset = [\out_{1},
  \dots, \out_{\nsamp}]$, $\out_{j}\in\{0, 1\}$ be the set of outcomes
or class labels e.g. $0\rightarrow$ healthy and $1\rightarrow$
faulty. Note that a time-series instance
$\ts_{i}\in\real^{\nts_{i}\times\nsens}$ is generated by an equipment unit over time and is associated with the equipment unit's label $\out_{i}$; also time-series instances $\ts_{i}$
can be of different lengths $\nts_{i}$, the dimensionality of the
time-series $\nsens$ is considered to be fixed across all instances.

\section{Our Approach}\label{sec:our-approach}

A typical textual document consists of a hierarchy of entities: characters,
words, phrases, sentences, paragraphs, sections, chapters. We propose a textual
transform to convert time-series data into a text document -- where individual
time stamped observations from each sensor can be thought of as
\emph{characters}, a sequence of characters desribing a temporal pattern will be
\emph{words}, and combination of such sequences will lead to \emph{phrases} or
\emph{sentences}. Observations from one sensor will form a \emph{section} of the
document, leading to a multiple section document for multivariate data. In this
way the entire time-series can be seen as an evolving \emph{journal} in a textual sense.

\begin{figure*}[t]
    \centering
    \includegraphics[width=\textwidth]{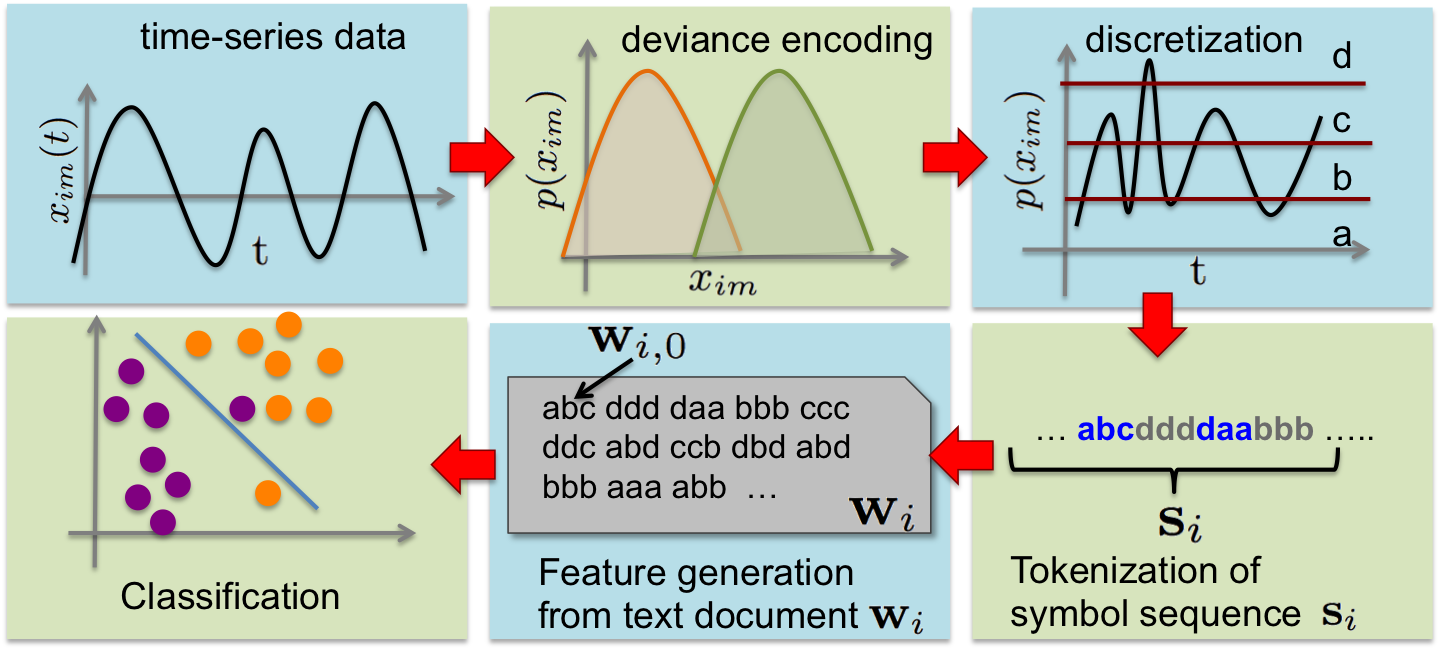}
    \caption{Process workflow for converting time-series to textual bag-of-words representation}
    \vspace{-12pt}
    \label{fig:workflow}
\end{figure*}

The next few sections describe the specifics of enabling the textual representation discussed above. The overall work flow is presented in Figure~\ref{fig:workflow}. 

\subsection{Encode abnormal behavior}\label{sec:encode_abnormal_behavior}
In this work, we intend to discover signatures that are indicative of degrading or damaged equipment. If the general expected behavior of a \emph{healthy} equipment is known a priori, then encoding the observations based on their alignment or deviation from the healthy state seems to be a reasonable choice. The healthy state operation of an equipment can either be derived from the physical or thermo-dynamical process governing the equipment, or it can be estimated in a completely data-driven way by acquiring historical records of operation in the healthy state. We choose the latter, more generic data-driven approach, since in many practical scenarios it may not be possible to completely describe a complex thermo-dynamical system with multiple interacting components in the context of changing and unknown or unrecorded environmental variables. Intuitively, we want to encode the observations by the likelihood of their occurrence with respect to a healthy unit. Formally, we need a map $f_1: \mathbb{R}^{\nsens} \rightarrow \mathbb{R}$ that converts the observed series into a series of likelihoods $\tuple{\mathbf{x}_i}_0^{T_i} \rightarrow \tuple{\Pr(\mathbf{x}_i)}_0^{T_i}$. Depending on desired complexity of modeling the likelihood, one may choose a simple unimodal distribution (such as Gaussian), a mixture model such as Dirichlet process based Gaussian mixture model (DPGMM)~\cite{antoniak1974}, or more complex models. In this work, we choose among four approaches: 

\begin{enumerate}
  \item \textbf{znorm-self}: The data is centered and
    scaled per unit using the mean and variance of data from \emph{each} unit. 
  \item \textbf{znorm-all}: The data is centered and scaled at the fleet-level using  the mean and variance computed using data from \emph{all} units.
  \item \textbf{DPGMM-univariate}: A uni-modal Gaussian distribution may not
    explain different modes of operation. A univariate DPGMM is inferred over observations from \emph{each} sensor across \emph{all} units. Then, each observation for a sensor is encoded as the likelihood of its occurrence using the DPGMM model.
  \item \textbf{DPGMM-multivariate}: A multivariate DPGMM is inferred over observations from \emph{all} sensors from \emph{all} units. Each multivariate observation is then encoded as the likelihood of its occurrence using the DPGMM.
\end{enumerate}





\subsection{Discretize and Symbolize}
Discretization of continuous features is a popular machine learning problem and several approaches have been studied in the past~\cite{Garcia:2013:SDT:2478560.2478975}. In this step, we map the real-valued series into a sequence of symbols. We define a mapping $f_2 : \mathbb{R} \rightarrow \Phi$ that is a partition on the space $\mathbb{R}$, and each cell of the partition maps to the appropriate symbol in $\Phi$. Thus the original series $(\obs_i)_0^{\nts_i}$ gets converted into a string $\mathbf{s}_i \in \Phi^{T_i}$, such that $\mathbf{s}_i = \tuple{f_2(\mathbf{x}_{i}(j)) : 0 \leq j\leq T_i}$, with length same as the original time series $|\mathbf{s}_i| = T_i$. 

We study two approaches for discretization: (a) \emph{Maximum Entropy Partition (MEP)}~\cite{Ray:2004:SDA:1013734.1013739} splits the data-space into $|\Phi|$ equiprobable regions. Popular for time series discretization~\cite{Apostolico:2002:MSL:565196.565200,Lin:2003:SRT:882082.882086}, it creates finer partitions in data-rich regions and coarser in partitions where data is less frequent in an unsupervised manner; (b) \emph{Recursive Minimal Entropy Partitioning (RMEP)}~\cite{conf/ijcai/FayyadI93} is a supervised approach to create partitions with minimum class entropy. Here we use encoded values from step 1~(section \ref{sec:encode_abnormal_behavior}) for all units, and label all data from a unit with it's final classification label. Using this dataset $\mathbb{D} = \{ \tuple{ x, y } : x \in \mathbf{x}_i, y = y_i, 1 \leq i \leq N \}$, we discover threshold $\theta$ that splits the instances into sets $\mathbb{D}_0$ and $\mathbb{D}_1$, such that class information entropy $H(\mathbb{D},\theta)$ is minimized. 
\begin{eqnarray*}
    H(\mathbb{D}, \theta) & = & \frac{|\mathbb{D}_0|}{|\mathbb{D}|}H(\mathbb{D}_0) + \frac{|\mathbb{D}_1|}{|\mathbb{D}|}H(\mathbb{D}_1) \\
    H(\mathbb{D}_j) & = & - \sum_{r \in \{0,1\}} \Pr(C_r,\mathbb{D}_j)\log\Pr(C_r,\mathbb{D}_j)\\
    \Pr(C_r, \mathbb{D}_j) & = & \frac{|\{\tuple{x,y} : \tuple{x,y} \in \mathbb{D}_j \wedge y = C_r\}|}{|\mathbb{D}_j|}
    \label{eqn:rmep}
\end{eqnarray*}
Threshold discovery is applied recursively using a stopping criteria based on the \emph{Minimal Description Length Principle}, thereby achieving multiple bins for discretization.

\subsection{Tokenize}\label{sec:tokenize}
Tokenization is the step of splitting long sequences into constituent tokens/words. Thus, we wish to split the symbolic sequence $\mathbf{s}_i$ into constituent tokens  $\mathbf{w}_i = \tuple{ \mathbf{w}_{i,0}, \ldots, \mathbf{w}_{i,K_i} }$, such that $ \mathbf{s}_i = \mathbf{w}_{i,0} \frown \ldots \frown \mathbf{w}_{i,K_i}$,  and $K_i \ll T_i$. , where $\frown$ is the string concatenation operator. In most natural languages, tokenization is enabled by punctuation or morphology. However, tokenizing a symbolized time-series is an open challenge. The simplest approach involves generating tokens of fixed lengths $L$, where token length $L$ is found  via cross-validation. 


As an alternative approach to determine $L$, we consider the symbol sequence as a Markov Chain over $|\Phi|$ symbols, with order $L$. The order is the number of past symbols that a future symbol depends on, or, $\Pr(\mathbf{s}_{i,m} | \mathbf{s}_{i,m-1}, \ldots, \mathbf{s}_{i,0}) = \Pr(\mathbf{s}_{i,m} | \mathbf{s}_{i,m-1},\ldots, \mathbf{s}_{i,m-L})$. We use the Conditional Mutual Information (CMI) based approach to estimate $L$~\cite{journals/corr/abs-1301-0148}. This method has been shown to be demonstrably better than previously well known approaches based on Bayesian information criterion (BIC), Akaike Information criterion (AIC), the Peres-Shields estimator or the ones based on the $\phi$-divergences~\cite{journals/corr/abs-1301-0148}. In the CMI-based approach the goal is to find the smallest $L$ such that 
\begin{equation}\nonumber
I_c(L+1) = I(\mathbf{s}_{i,m} ; \mathbf{s}_{i ,m-L} | \mathbf{s}_{i,m-1},\ldots, \mathbf{s}_{i,m-L+1}) = 0
\end{equation}
where $I_c(L+1)$ is the conditional mutual information between observations that are $L$ time steps apart. Once estimated, $L$ is used as the token-length. 
\subsection{Generate features}
With the tokenization step, we have transformed the original time-series into a sequence of \emph{punctuated} symbol sequences akin to words in a text document. This representation lets us leverage the popular \emph{bag-of-words} representation that is successful in the text-mining domain. In this approach, each document is represented as a vector of term weights. The vector magnitudes represent a score for every word in the document. For our model, we use the popular TF-IDF (Term Frequency - Inverted Document Frequency) score, specifically the variant commonly known as \emph{ltc}. The vocabulary is defined as the union of all the tokens resulting from all the time-series, $\mathbb{V} = \{ w | w \in \mathbf{w}_i, 1 \leq i \leq N \}$. If $\#(j | \mathbf{w})$ denotes the number of occurrences of the token $j$ in the \emph{textual} transform $\mathbf{w}_i$ of time-series $(\obs_i)_0^{\nts_i}$, then the \emph{ltc} scores can be calculated as follows: 
\begin{eqnarray*}\label{eqn:tfidf}
    \mathbf{w}_i & \rightarrow & \mathbf{b}_i := \frac{[b_1,\ldots,b_{|\mathbb{V}|}]}{\|[b_1,\ldots,b_{|\mathbb{V}|}]\|}   \\
    b_j & = & \log(\#(j | \mathbf{w}_i) + 1) \log\frac{|N|}{|\{ \mathbf{w_{k}} : \exists j \in \mathbf{w_{k}}, \mathbf{w_{k}} \in \{\mathbf{w}_1,\ldots, \mathbf{w}_N \} \}|}
\end{eqnarray*}
where each $b_j$ is the score for a token in the textual transform

\subsection{Classify}
Given the bag-of-words representation achieved in the previous step, now it is straightforward to apply any popular text-classification algorithm for learning a model that distinguishes between healthy and unhealthy units. We choose the linear kernel based Support Vector Machine (SVM), a very popular and successful model for text classification.

\subsection{Computational Complexity}\label{sec:complexity}
In Table~\ref{tab:complexity}, we list the computational complexity of the major components of the textual transform process. The transformation procedure, in its simplest form, by choosing the fastest approach at each step, is linearly dependent on the size of the dataset, making it highly practical for large-scale deployment. Since the steps involve mere iterations over the dataset, this is highly amenable to big-data systems based on map-reduce that are highly popular among industrial monitoring systems. Even with the most expensive choice of approach at each step, the algorithm does not suffer significantly since it is still linear in the number of instances and the length of the time-series, possibly the largest variables for practical prognostics deployment. 



\begin{table*}[t]
    \centering
    \begin{tabular}{lll}
        \toprule
        Encode & Discretize & Tokenize \\
        \midrule 
        Gaussian:  $\mathcal{O}(NMT)$ & MEP: $\mathcal{O}(NMT)$ & Fixed-length: $\mathcal{O}(NMT)$ \\
        DP-GMM~\cite{Blei05variationalinference}: $\begin{cases} & \text{(covariance)} \\ \mathcal{O}(NMT), & \text{diagonal} \\ \mathcal{O}(NM^2T), & \text{tied} \\ \mathcal{O}(NM^3T), & \text{full}\end{cases}$ & RMEP:$ \mathcal{O}(MNT\log(NT))$~\cite{kohavisahamidiscretize} & Markov Order: $\mathcal{O}(NMT)$ \\
        \bottomrule
    \end{tabular}
    \caption{Computational Complexity of the major steps in the textual transformation process. $T := \text{max}_i T_i$. Typically, $M \ll N$ and $M \ll T$. For avoiding notation clutter, we have ignored several quantities that linearly (or sublinearly) affect the complexities, but are insignificant in magnitude compared to $N,M,T$. These include, among others, the number of mixture components for DPGMM, the number of splits for RMEP, or the number of candidate lengths tested for Markov Chain order estimation.}
    \label{tab:complexity}
\end{table*}

\section{Data sets}\label{sec: dataset}
The proposed methods have been deployed on proprietary datasets from commercial aircraft
engines. To maintain confidentiality, while still retaining reproducibility of the results, the
experiments in this paper are presented on 4 similar public benchmark datasets and simulated
datasets with multiple levels of signal to noise ratio.

\subsection{Real aircraft engine dataset}\label{subsec: ge dataset} This proprietary dataset
consists of data from $59$ aircraft engines of which $20$ engines are distressed or known to
have usage-based equipment underperformance. The other $39$ engines are known to be in a
healthy state. The determination of healthy and distressed engines is based on detailed manual
inspections that are cumbersome and time consuming, hence the need for automatic detection of distressed engines. 
Each flight, or cycle of operation, of an engine generates a row of
data for that engine detailing the sensor values measured at certain critical phases of
operation such as flight takeoff. We used multivariate measurements from 10 chosen sensors 
for $18$ months of usage for each of the chosen engines. In this timeframe, each engine has flown about $650-1560$ times, leading to a dataset
with approximately half a million sensor observations.

\subsection{C-MAPSS Turbofan dataset}\label{subsec: cmapss dataset}
These datasets, created and publicly
released by the NASA Prognostics Center of Excellence, consist of
run-to-failure observations of aircraft gas turbine engines using the C-MAPSS tool-kit~\cite{CMAPSS}.

\begin{figure}[h]
  \centering
  \includegraphics[width=0.8\columnwidth]{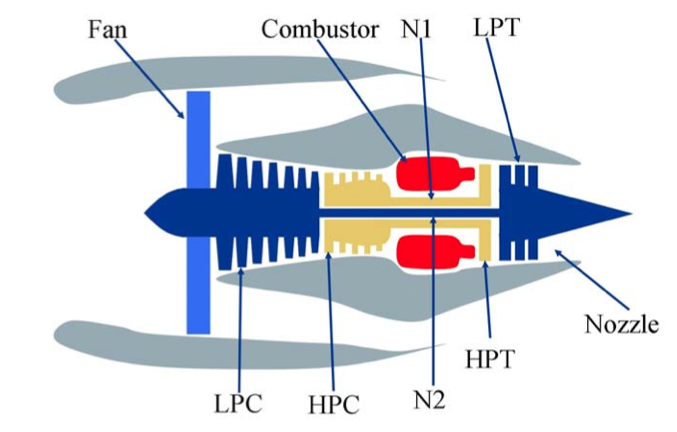}
  \caption{Schematic of a Turbofan engine~\cite{CMAPSS}}
  \vspace{-12pt}
  \label{fig:engine}
\end{figure}

The gas turbine engine system consists of five major rotating
components -- fan (F), low pressure compressor (LPC), high pressure
compressor (HPC), high pressure turbine (HPT), and low pressure
turbine (LPT), as seen in Figure~\ref{fig:engine}. Deterioration $\fault$ of
the engine is modeled as a loss of flow and loss of
efficiency of the HPC module. Starting from a randomly chosen initial
state, deterioration rate $\rate$ is modeled as an
exponential rate of change for both modes of failures.

Four datasets are available, with data for 100, 100, 249 and 260
engines in each set respectively. Each dataset corresponds to a
certain mode of damage propagation under different conditions. For
each engine, 21 sensor measurements and 3 operating parameters are
recorded for every cycle of engine operation. Six different operating
conditions determined by altitude, Mach number and throttle resolver
angle are present. Lifespans of engines are in the range 128-525
cycles.

\begin{figure}[h]
   \centering
   \includegraphics[width=\linewidth]{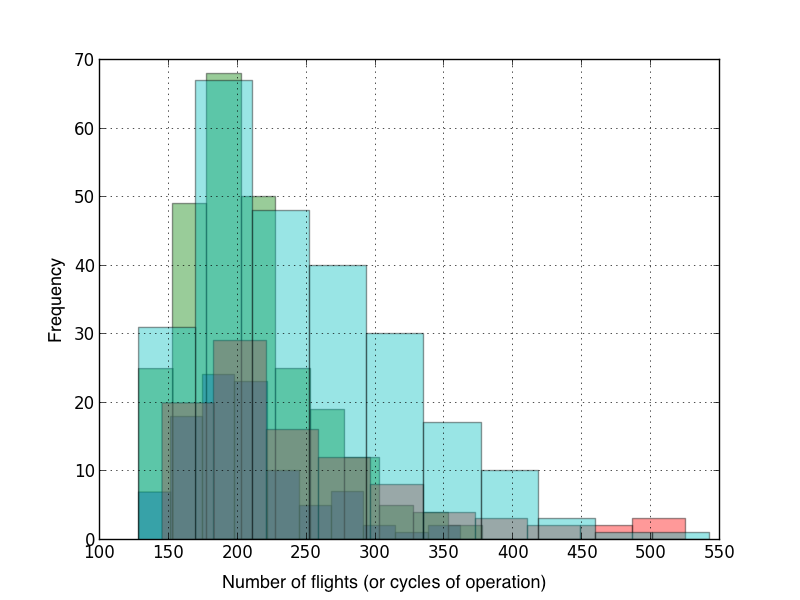}
   \caption{Distribution of the life expectancies of units in the four C-MAPSS datasets}
   \label{fig:cmapss_life_expectancy}
   \vspace{-12pt}
 \end{figure}

Figure~\ref{fig:cmapss_life_expectancy} depicts the distribution of life-expectancies of the units, in terms of number of flights, in the four datasets.

We use these datasets for the task of identifying engines that have
lower life expectancy. We have split engines in each datasets into two
classes based on whether they have higher or lower than median
lifespan in that dataset, leading to roughly balanced classes.  This
is a realistic industrial setting where identification of units with
lower than usual life expectancy allows timely root-cause analysis,
increased monitoring, preventive maintenance and operational
adaptations. As the detection of under-performance needs to happen
early in life, we use observations only from the initial 50 cycles of
operation for training and testing splits.

\subsection{Duffing oscillator dataset}\label{subsec: duffing dataset}
\begin{figure*}[!htb]
\begin{center}
	\begin{tabular}{ccc}
	\includegraphics[width = 0.3\textwidth]{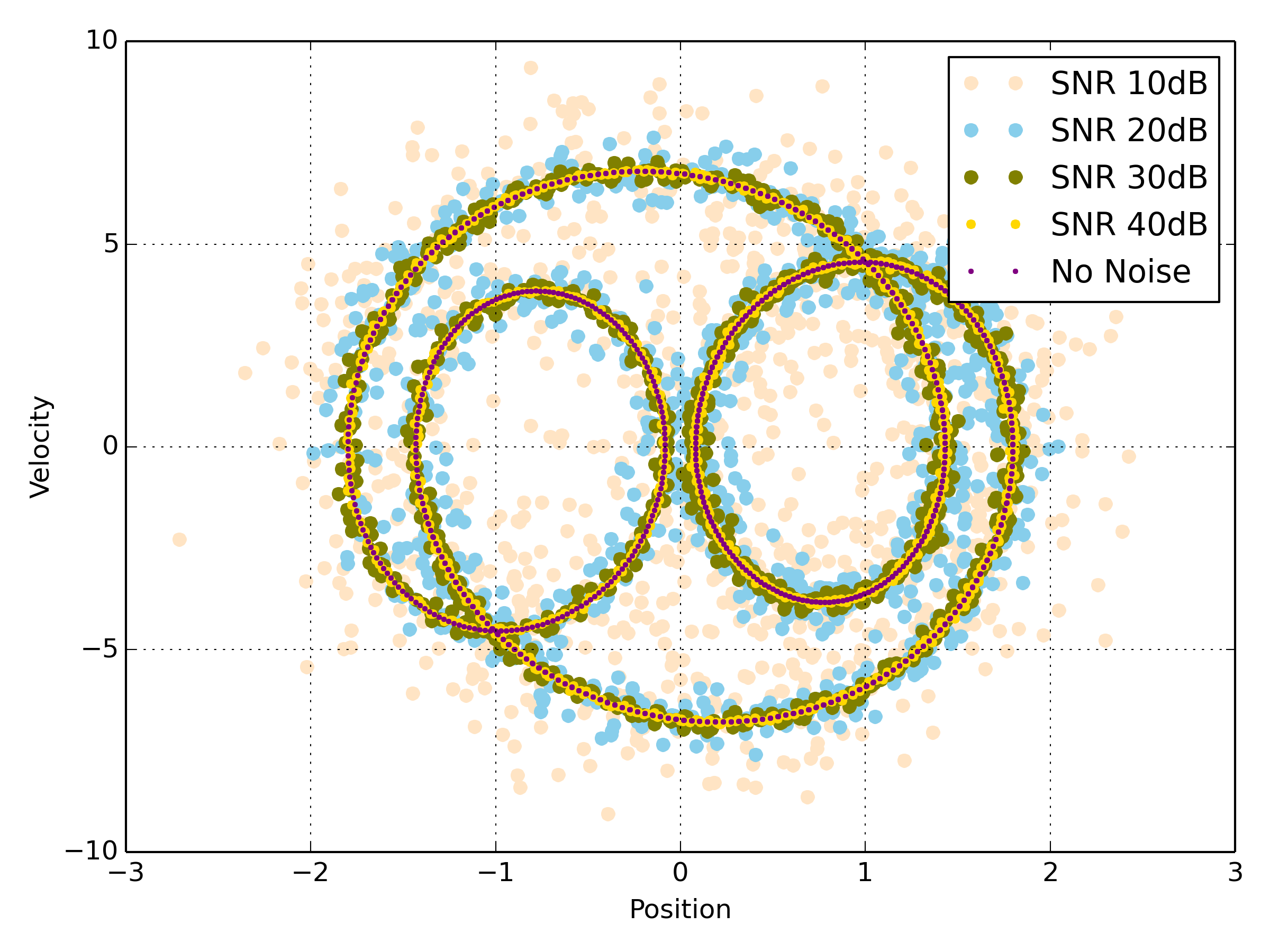}
	&
	\includegraphics[width = 0.3\textwidth]{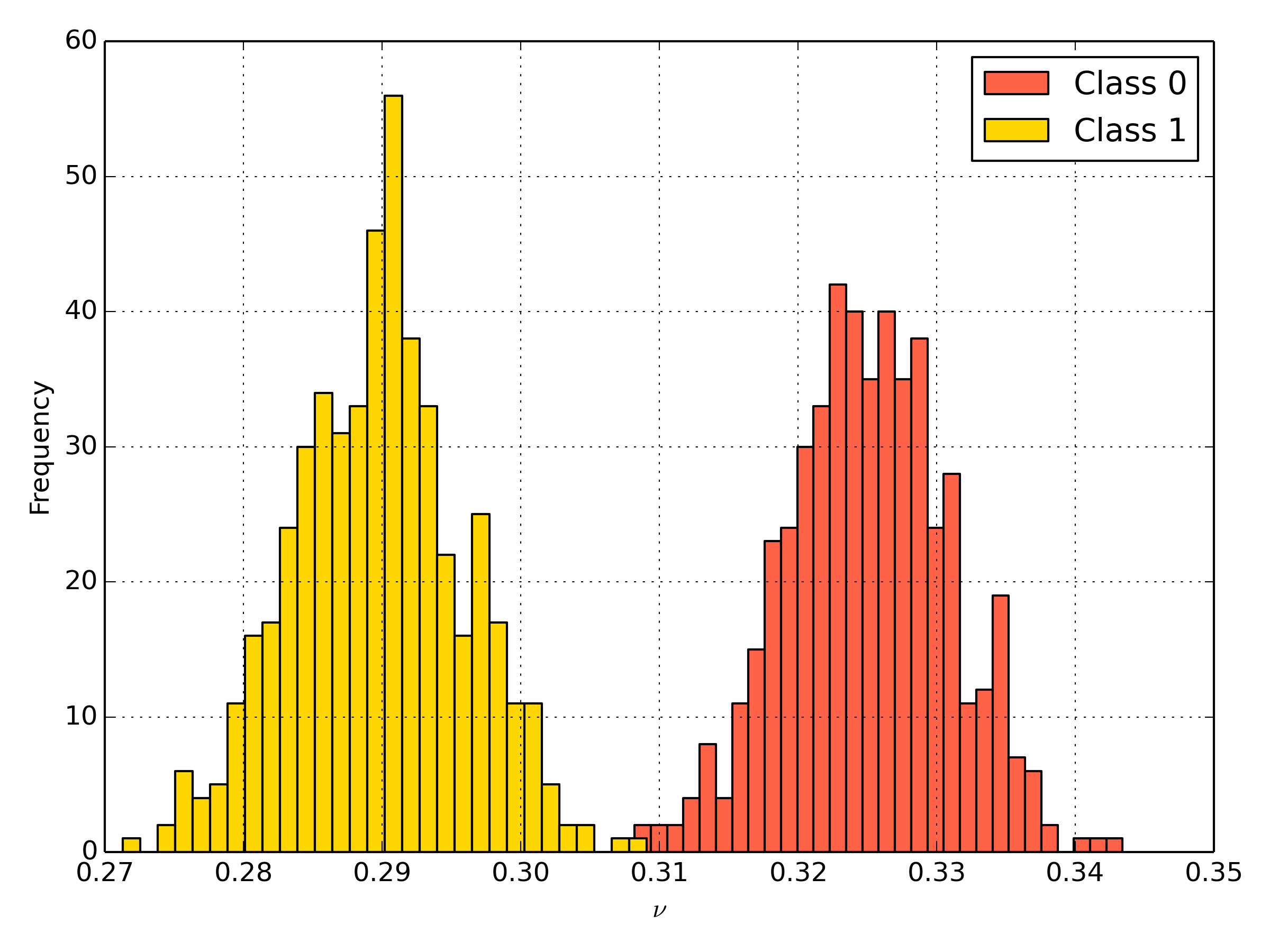}
	&
	\includegraphics[width = 0.3\textwidth]{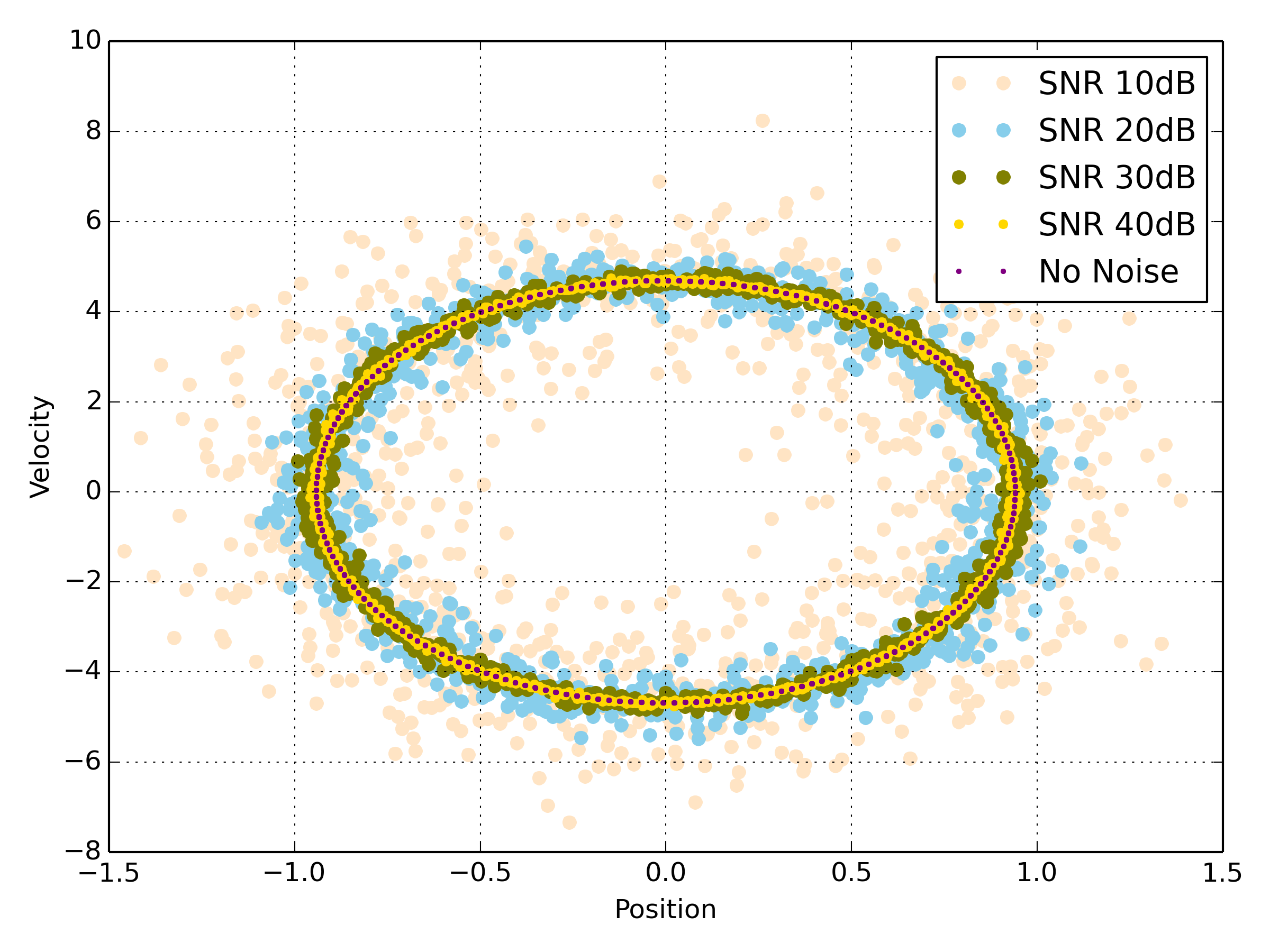}
	\\
	(a)  & (b)  & (c)
	\end{tabular}
	\caption{Phase plot of the duffing oscillator for various noise levels (a) $\nu(t_{s}) = 0.29$, (b) $\nu$ distribution used for creating two classes, and (c) phase plot for $\nu(t_{s})=0.325$}%
	\label{fig:nu300_duffing_phase_plots}%
	\vspace{-20pt}
\end{center}
\end{figure*}

Another set of data is generated using the Duffing Oscillator which is
described by the following second order non-autonomous system:
\begin{equation}\label{eq:duffingEq}
\frac{d^2x(t)}{dt^2} + \nu(t_{s})\frac{dx(t)}{dt} + \kappa x(t) + \alpha x^3(t) = A \cos(\Omega_f t) 
\end{equation}
The system dynamics evolve in the time-scale of the time parameter $t$
while the quasi-static dissipation parameter $\nu(t_{s})$ changes
slowly in time $t_{s}$. This system describes the motion of a damped
oscillator across a twin well potential function. This equation has
been used as the underlying model for many engineering systems such as
those for electronic circuits, structural vibrations and chemical
bonds~\cite{B97} . This system shows a slow change in dynamics as the
parameter $\nu(t_{s})$ is increased with an abrupt change of behavior
or a {\it phase change} for $\nu(t_{s}) >0.3$. This behavior is
analogous to fault propagation in engineering systems where slow
changes in system parameters such as component efficiency or corrosive
damage leads them from nominal to anomalous and finally to an abrupt
failed state. The parameter $\nu(t_{s})$ can be thought of as the {\it
  health} parameter which is typically unobserved or not measured. The
goal is to infer the health state of the system using only its
observations -- position and velocity in this case.

Duffing data for classification experiments, is generated
by solving the Duffing equation for different values of $\nu(t_{s})$
which are sampled from a bimodal Gaussian distribution with means
$0.290$ (healthy) and $0.325$ (faulty) and a common standard deviation
of $0.00625$; parameter $A=22.0$, $\Omega_{f} = 5.0$, $\kappa = 1.0$
and $\alpha= 1.0$ with an initial condition of $\trans{[0.0,
    \ 0.1]}$. The {\tt scipy} package for {\tt Python} is used to
solve equation~\ref{eq:duffingEq} using runge-kutta method of order
(4)5 using a time-step of $\Delta_{t}=T_{f}/200$, and then
down-sampled to create $\approx200$ time-steps long time-series data
for position and velocity for each value of $\nu(t_{s})$; here $T_{f}
= 2*\pi/\Omega_{f}$ is the time-period of the forcing function. Four
data-sets consisting of $500$ samples each of healthy and faulty are
used is created with four levels of signal-to-noise ratio (SNR) -- 10dB, 20 dB, 30 dB, 40 dB and clean.

\section{Experiments}\label{sec:experiments}
\subsection{Experimental Setup}\label{subsec:exp-setup}
We developed our system in Python and used the scikit-learn\footnote{http://scikit-learn.org/}
library of algorithms. All the results reported have been estimated using nested cross-validation. That means, the outer cross-validation averages performances over multiple training-testing splits, and the inner cross-validation is done for model-selection/parameter-tuning over multiple training-validation splits. For the inner model-selection loop and outer accuracy estimation, we perform 10-fold cross-validation~\cite{Kohavi:1995:SCB:1643031.1643047}. For the support vector machines we chose the hinge-loss or margin-penalty parameter, commonly denoted as $C$, from the set $\{10^{-2},\ldots,10^{2}\}$.


\subsection{Classification Experiments}\label{subsec:exp-classification}
We use both non-temporal and temporal state-of-the-art algorithms as baselines for comparison with the proposed method:

\begin{itemize}
\item \textbf{Non-temporal}: In this setting, we use data from last $k$ cycles and label them with the final classification label of each unit, leading to $k$ labeled instances. We use this data to train an SVM and call this approach SVM-final (\textbf{SVM-F}). In addition to the final instances, we use the initial few cycles from each unit and label those as the healthy class, as all units were healthy in the beginning. We call this approach SVM initial-final (\textbf{SVM-IF}). We have also tried using all the data from each unit and labeling them with the final classification label, but the performance was significantly worse as all units are healthy initially. We also use RandomForest classifiers instead of SVM to get additional non-temporal baselines \textbf{RF-F} and \textbf{RF-IF}. The parameter $k$ was tuned via cross-validation. 
\item \textbf{Temporal}: Here we treat the problem as time-series classification (TSC) and use 1-Nearest Neighbor classifier with DTWCV -- Dynamic Time Warping metric with wraping window set by cross-validation, for classification. Our choice is motivated by a comprehensive evaluation of TSC algorithms that says that ``a new algorithm for TSC is only of interest to the data mining community if it can significantly outperform DTWCV"~\cite{DBLP:dblp_conf/sdm/LinesB14}. We call this approach NNDTW.
\end{itemize}



Our approach called TTC (for Textual Time-series enCoding) and its variants
are summarized in Table~\ref{tab:ttc-acronyms}

\begin{table}[h]
  \centering
  \begin{tabular}{ lll}
    Deviance Encoding & Discretization & Tokenizaton \\
    \toprule
    S: znorm-self & M: MEP  & E: Equilength  \\
    A: znorm-all  & R: RMEP  & M: Markov \\
    U: DP-GMM univariate  & & \\
    M: DP-GMM multivariate  & & \\
    \bottomrule
  \end{tabular}
  \caption{Quick reference for the naming convention used on the TTC
  configurations. Each configuration is a 3-letter acronym based on the
  strategy utilized for each of the three steps. For example, TTC-SME means the znorm-self(S) was used for
encoding deviance, MEP(M) was used for discretization and Equilength(E) tokens
were used for the tokenization step}
  \label{tab:ttc-acronyms}
\end{table}

Table~\ref{tab:classresults} provides the results of the
classification experiments. The textual-transform based models, TTC-*,
clearly outperform other approaches with a significant margin. The
superior performance over non-temporal approaches indicate that there
is much to be gained by utilizing the temporal nature of the
observations from each unit, especially in the case of C-MAPSS dataset
and the noisy versions of the Duffing dataset. 

The performance of the strong TSC baseline, DTWCV, is particularly
weak, suggesting that this problem is not a typical TSC problem. Since
engines are operated independently, controlled differently, behave in
idiosyncratic ways, and observations are typically noisy, there is not
much to be gained by directly comparing the \emph{shapes} of their
time-series profiles, a common underlying theme of TSC algorithms,
particularly of the DTWCV. A unique aspect of sensor time-series from
engineering systems is fixed rate sampling for data collection. In
this scenario, warping the time-axis for DTW calculation can lead to creation of artifacts
not intrinsic to the original shape or pattern. This might also
explain the poor performance of the proven baseline NNDTW in this
study. The performance of DTWCV improves on the duffing dataset as the
signal to noise ratio (SNR) improves. However, practical scenarios rarely
lead to clean observations.

On the other hand, TTC approaches benefit from aggregating the common
occurrences through the bag-of-words representation and thereby
simulating a \emph{cumulative damage} model of the engine. Thus, if an
engine has a high occurrence of mishandling, or rough weather, the
model will be able to aggregate those through term-frequency and weigh
them appropriately using the frequency of their occurrence through the
population (inverse document frequency). The SVM classifier at the end
of the TTC steps would then weigh the damage precursors highly to learn a superior model for prognostics.

\begin{table*}\label{table: xxx}
\begin{center}
\begin{tabular}{l|c|c|c|c|c|c|c|c|c|c|c}
     \multirow{2}{*}{\textbf{Method}} & \multicolumn{1}{c|}{Jet Engines} & \multicolumn{4}{c|}{C-MAPSS Dataset} & \multicolumn{5}{c|}{Duffing Dataset} \\
  \cmidrule(r){2-11}
  & 1 & 1 & 2 & 3 & 4 & 10dB & 20dB & 30dB & 40dB & clean \\
     \midrule
\hline
 MF & 0.660 & 0.506 & 0.506 & 0.506 & 0.500 & 0.500 & 0.500 & 0.500 & 0.500 & 0.500\\
 RF-F & 0.730 & 0.478 & 0.533 & 0.663 & 0.671 & 0.675 & 0.790 & 0.864 & 0.981 & 0.998\\
 RF-IF & 0.710 & 0.536 & 0.506 & 0.506 & 0.565 & 0.621 & 0.809 & 0.859 & 0.980 & 0.999\\
 SVM-F & 0.660 & 0.528 & 0.533 & 0.506 & 0.654 & 0.638 & 0.819 & 0.823 & 0.819 & 0.831\\
 SVM-IF & 0.560 & 0.528 & 0.486 & 0.506 & 0.626 & 0.621 & 0.712 & 0.711 & 0.819 & 0.831\\
 NNDTW & 0.730 & 0.526 & 0.568 & 0.523 & 0.500 & 0.752 & 0.839 & 0.987 & 0.999 & 0.999\\
 TTC-SME & 0.793$^\bigstar$ & 0.637 & 0.618 & 0.606 & 0.455 & 0.892 & 0.992$^\bigstar$ & 0.995 & 0.998 & 0.997\\
 TTC-SMM & 0.793$^\bigstar$ & 0.688$^\bigstar$ & 0.525 & 0.667 & 0.473 & 0.852 & 0.983 & 0.998 & 0.998 & 0.999\\
 TTC-SRE & 0.727 & 0.546 & 0.495 & 0.564 & 0.516 & 0.852 & 0.984 & 0.993 & 0.999 & 0.999\\
 TTC-SRM & 0.677 & 0.574 & 0.494 & 0.614 & 0.500 & 0.841 & 0.983 & 0.991 & 0.997 & 0.997\\
 TTC-AME & 0.710 & 0.578 & 0.634$^\bigstar$ & 0.726 & 0.714$^\bigstar$ & 0.894$^\bigstar$ & 0.985 & 0.999$^\bigstar$ & 1.000$^\bigstar$ & 0.998\\
 TTC-AMM & 0.740 & 0.619 & 0.529 & 0.726 & 0.553 & 0.861 & 0.989 & 0.999$^\bigstar$ & 1.000$^\bigstar$ & 0.998\\
 TTC-ARE & 0.593 & 0.537 & 0.595 & 0.727$^\bigstar$ & 0.674 & 0.883 & 0.987 & 0.993 & 0.998 & 0.999\\
 TTC-ARM & 0.660 & 0.578 & 0.617 & 0.727$^\bigstar$ & 0.600 & 0.882 & 0.985 & 0.993 & 0.998 & 0.999\\
 TTC-UME & 0.727 & 0.567 & 0.572 & 0.618 & 0.502 & 0.825 & 0.949 & 0.997 & 0.996 & 0.999\\
 TTC-UMM & 0.760 & 0.423 & 0.479 & 0.426 & 0.550 & 0.837 & 0.959 & 0.998 & 0.999 & 0.999\\
 TTC-URE & 0.707 & 0.504 & 0.603 & 0.526 & 0.516 & 0.726 & 0.881 & 0.987 & 0.984 & 0.999\\
 TTC-URM & 0.657 & 0.524 & 0.482 & 0.507 & 0.530 & 0.589 & 0.828 & 0.991 & 0.937 & 0.999\\
 TTC-MME & 0.643 & 0.506 & 0.533 & 0.506 & 0.465 & 0.807 & 0.886 & 0.980 & 0.999 & 0.997\\
 TTC-MMM & 0.710 & 0.506 & 0.478 & 0.506 & 0.500 & 0.764 & 0.840 & 0.979 & 0.998 & 1.000$^\bigstar$\\
 TTC-MRE & 0.543 & 0.494 & 0.537 & 0.494 & 0.470 & 0.754 & 0.819 & 0.886 & 0.998 & 0.966\\
 TTC-MRM & 0.443 & 0.494 & 0.482 & 0.494 & 0.500 & 0.500 & 0.503 & 0.760 & 0.959 & 0.765\\
\hline
 TTC-overall & 0.793 & 0.666 & 0.595 & 0.737 & 0.726 & 0.869 & 0.983 & 0.997 & 0.996 & 0.997\\
\hline
\end{tabular}
\caption{Accuracy of the classification approaches. The best performance for each dataset is
indicated by a $^\bigstar$. Refer to Table~\ref{tab:ttc-acronyms} for acronyms of TTC-variants. TTC-overall is an all-encompassing approach that is tuned over all the TTC-* approaches using only the training set to select the best one and then predict over the test set, for each train-test split}
\vspace{-12pt}
\label{tab:classresults}
\end{center}
\end{table*}


\subsection{Early detection}\label{subsec:early-detection}
\begin{figure}[h]
  \centering
  \includegraphics[width=\linewidth]{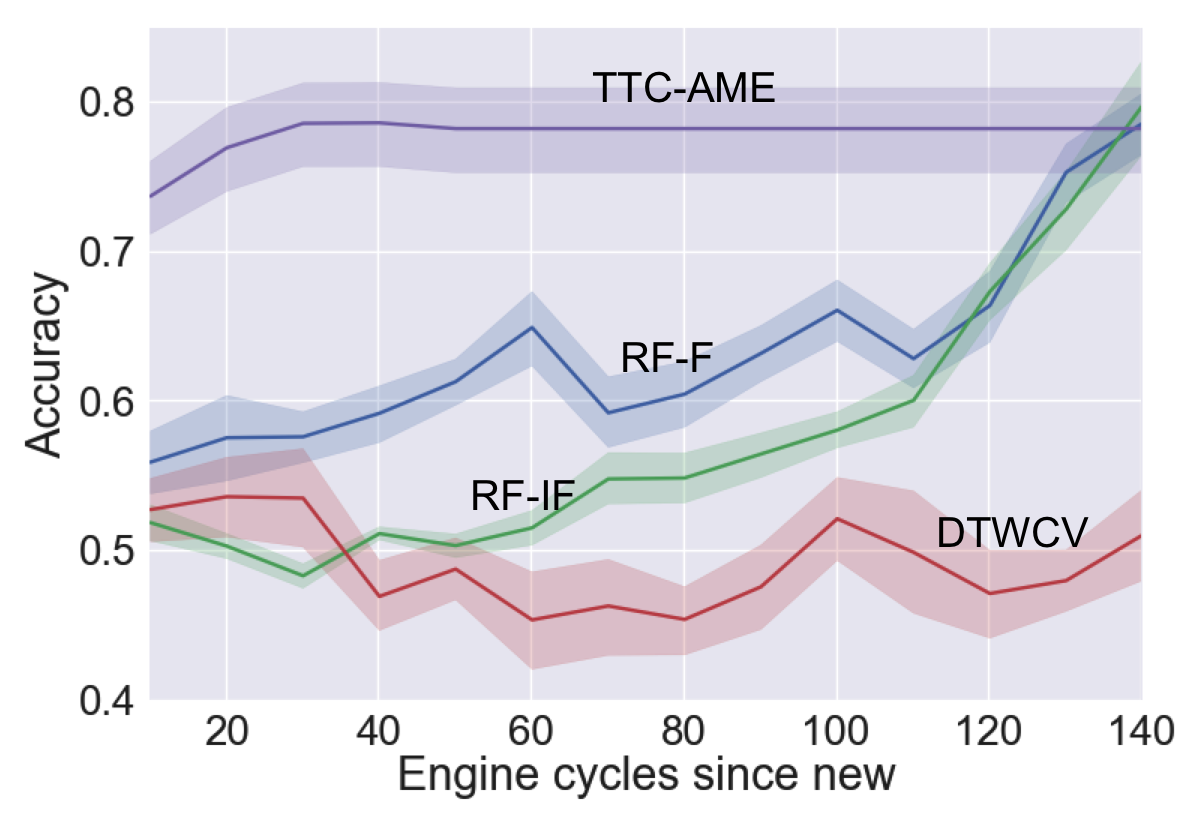}
  \caption{Comparison of the trend of accuracy as a function of the number of initial observations from a unit on the dataset C-MAPSS 4}
  \vspace{-12pt}
  \label{fig:trend}
\end{figure}

In this section, we compare the performance of candidate algorithms in
detect units with low life expectancy early in their life. For this
experiment, we trained all algorithms using initial $150$ cycles of
units in the training set, and evaluated prediction accuracy of the
trained model on the test set as a function of increasing number of
cycles observed from the test units. Figure~\ref{fig:trend} depicts
the performance trend as a function of the increasing number of cycles
for the test set. The results have been averaged over 10-fold
cross-validation. It can be observed that our approach is
significantly better than the baselines in the initial stages. The
baselines, especially RF-F and RF-IF, achieve comparable performance
as more information from the test unit becomes available. However, it
is too late.


\subsection{Analyzing the textual transform}\label{subsec:tt-analysis}

\begin{figure}[h]
\centering
\includegraphics[width=\columnwidth]{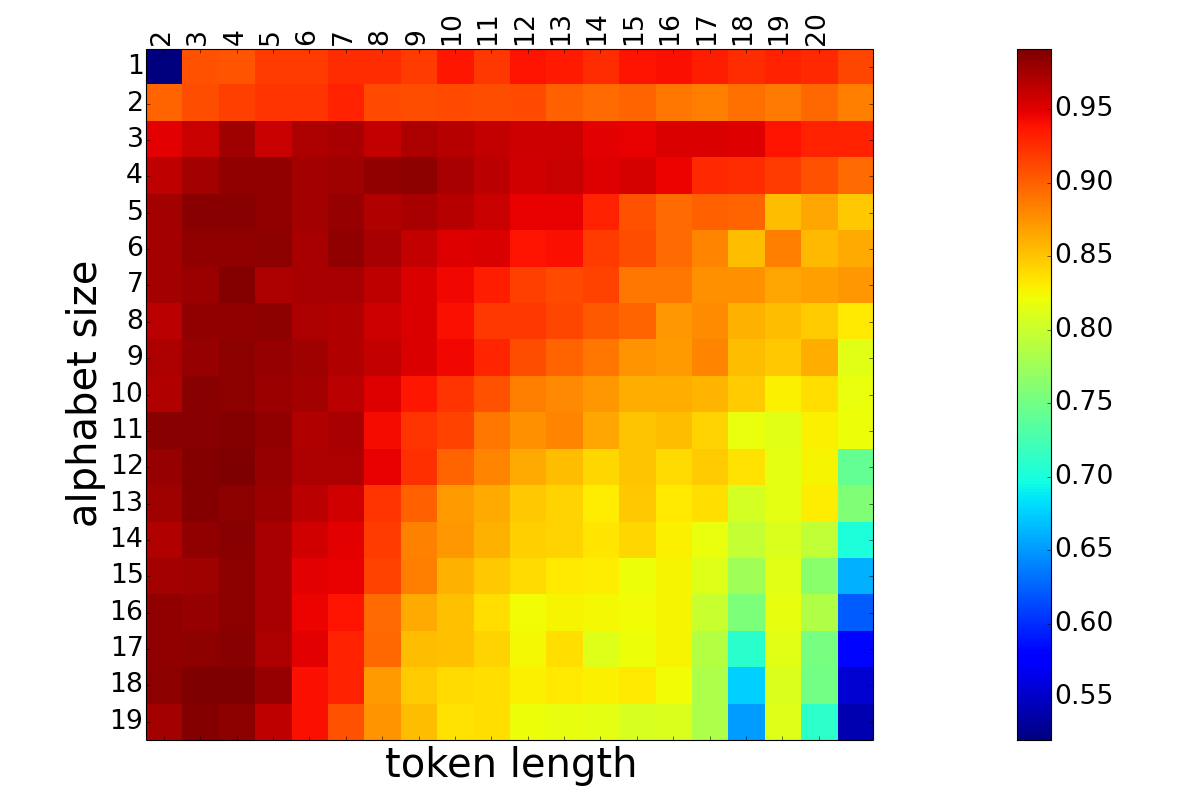}
\caption{Effect of word-length $L$ and alphabet-size $|\Phi|$ on classification accuracy for dataset Duffing SNR 20dB. Colors denote levels of accuracy, see colorbar}\label{fig:asize-wordlen-comparison}
\end{figure}

Words in natural language documents have been commonly desrcibed by the power-law distribution. Being long-tailed, it has a few very common \emph{stop words}, and many infrequent \emph{rare words}. The stop-words, being common to all documents, do not provide distinguishability, while rare words do not generalize to larger corpus. Consequently, the \emph{middle-class} of the power-law distribution is critical to good predictive performance. In our problem, few discrete symbols ($1 \leq |\Phi| \leq 3$), with short word lengths ($L \leq 3$) leads to too many stop-words, since shorter sequences of small number of symbols are likely to be observed very commonly in the time-series, thereby leading to low discrimination among different behaviors. On the other hand, many discrete symbols ($|\Phi| \geq 15$), with long word-lengths ($L \geq 10$), leads to many rare words, thereby making it infeasible to get good generalizable performance. This trend of accuracy with respect to the symbol size and token length is evident in the Figure~\ref{fig:asize-wordlen-comparison}. From the Figure~\ref{fig:asize-wordlen-comparison}, it can be observed that the optimal performance occurs at alphabet-size $|\Phi| = 10$ and token length $L = 4$ and the performance tapers down around this.

As an alternative approach for determining token length, we had suggested that the symbolic time-series may be modeled as a Markov Chain and its order may be estimated.  From our experiments, we observe that the order of the Markov Chain representing the symbolic time-series comprising of 10 symbols is about 4 for the Duffing dataset with SNR 20dB. Thus, every symbol sequence of 4 consecutive observations contains sufficient history that's predictive of the future symbol to come. This aligns well with our empirical assessment that the optimal word-length for symbol size 10 is indeed around 4, as seen in Figure~\ref{fig:asize-wordlen-comparison}. 

\subsection{Interpretability}\label{sec:interpretability}
\begin{figure*}
\begin{center}
  \begin{tabular}{cc}
    \includegraphics[width=0.45\textwidth]{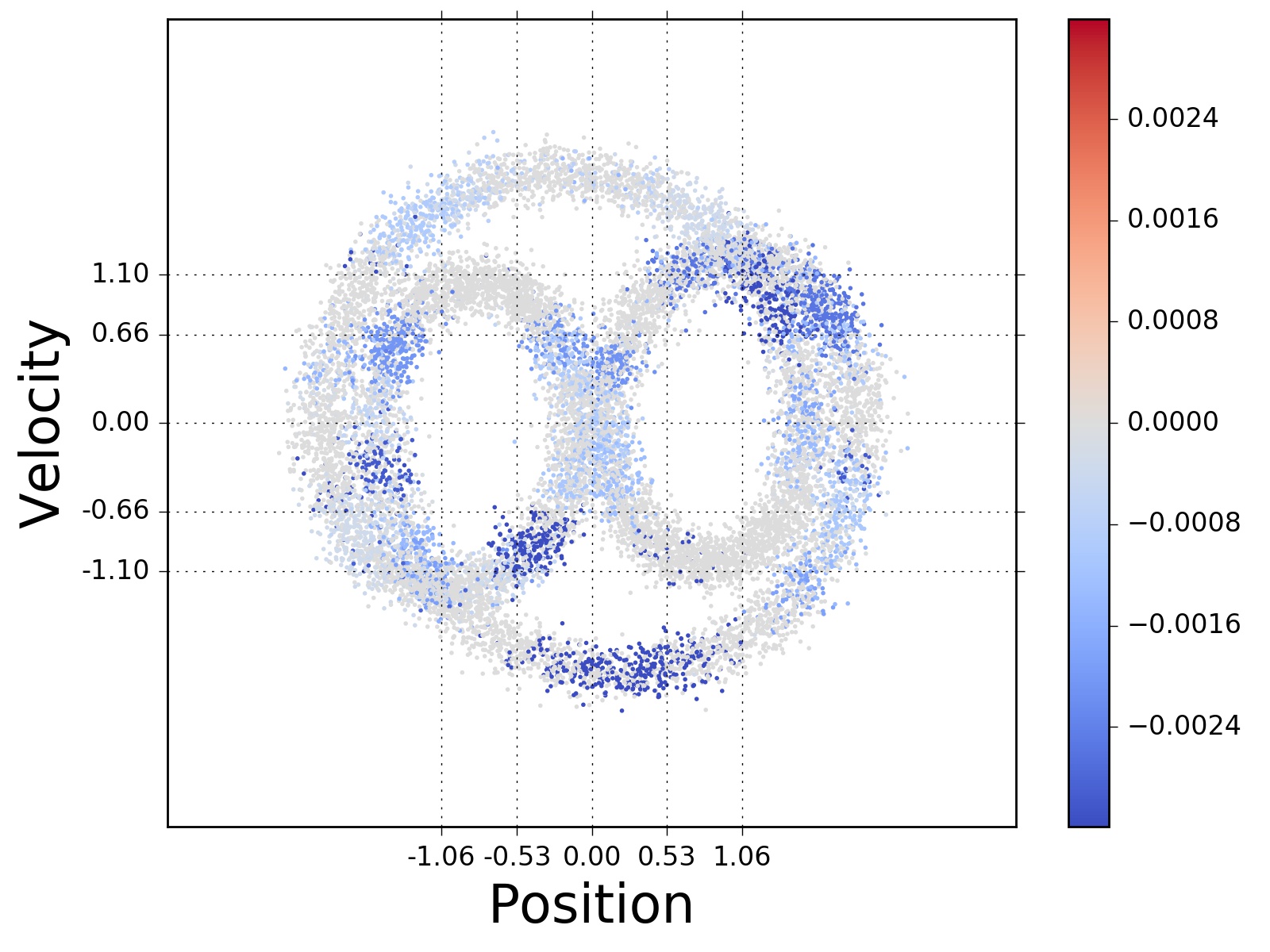}
    &
    \includegraphics[width=0.45\textwidth]{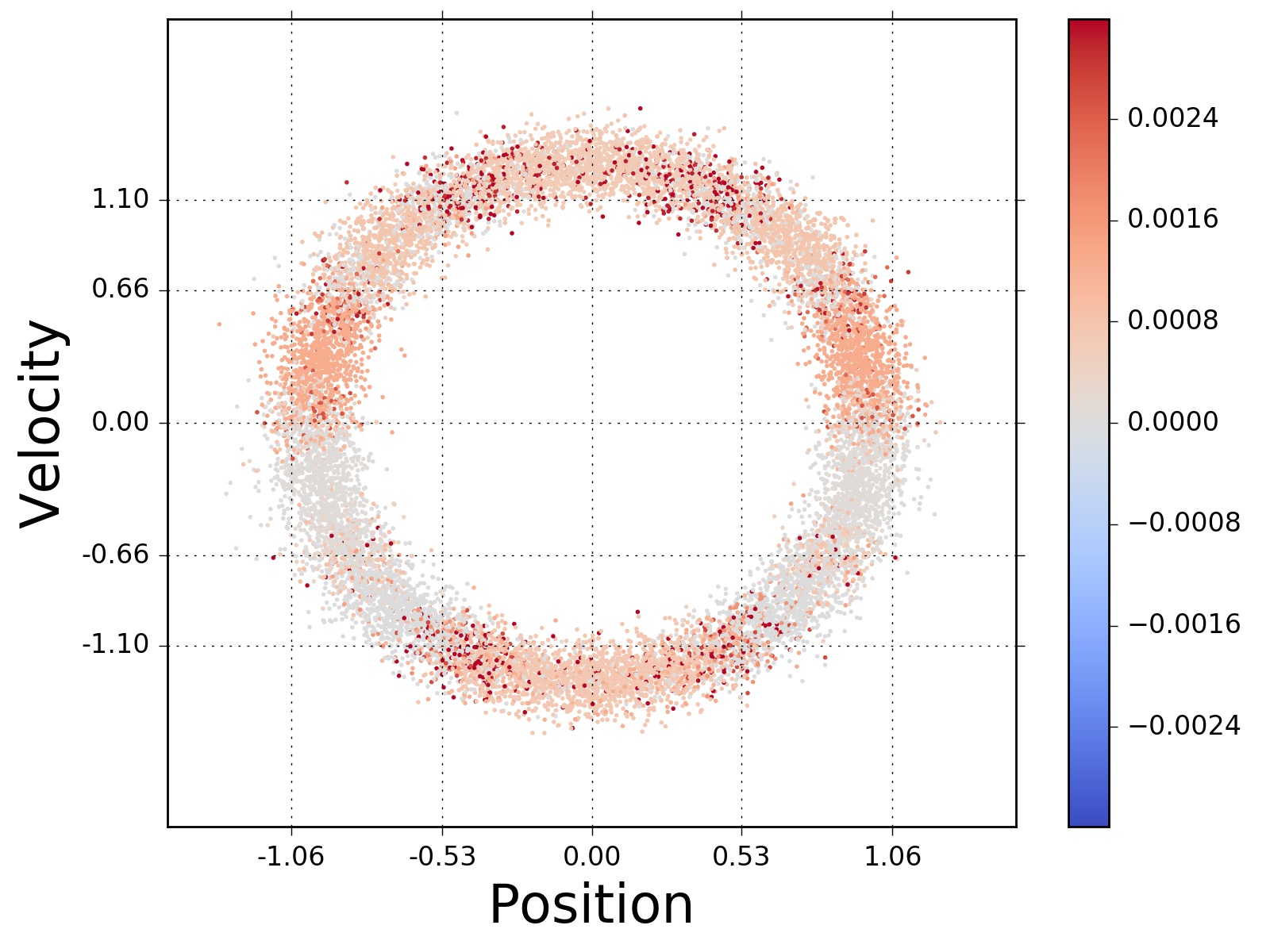}
    \\
    (a) & (b)
  \end{tabular}
  \caption{Feature weights of words formed using velocity sensor data for Duffing dataset with SNR 20dB shown as color map; weights are shown for the AME variant of proposed TTC approach. Subfigure (a) shows large negative weights for healthy or the negative class, and subfigure (b) shows large positive feature weights for the faulty or the positive class; critical segments such as transition points into inner loop formation in the negative class are correctly selected by the approach}\label{fig:interpretability}
  \vspace{-12pt}
\end{center}
\end{figure*}

Another aspect of the proposed approach is the interpretability of results. Using feature weights it is possible to identify the words (or features) that were most useful for discriminating between the two classes. These words, or sequences of symbols, can be mapped back to the original dataset, thus identifying time-series subsequences that were important for classification. Figure~\ref{fig:interpretability} shows as a colormap the weights for important words identified by TTC for classification of healthy and faulty classes; feature weights depicted are for words constructed from the symbols of velocity data. It can be seen in subfigure (a) that for healthy class  (the negative class), important segments with large negative weight are around the transition points -- locations where the phase-space plot diverges away to form the inner loops. Similarly, subfigure (b) shows for the faulty class (the positive class) the time segments associated with important words with large positive weights. With a similar approach, for the case of real aircraft engine, we were able to discover words that when mapped back to the time-series data helped the domain experts understand the earliest signatures of impending distress. 


\subsection{Identifying groups of units to monitor}
\begin{figure}[h]
  \centering
  \includegraphics[width=\columnwidth]{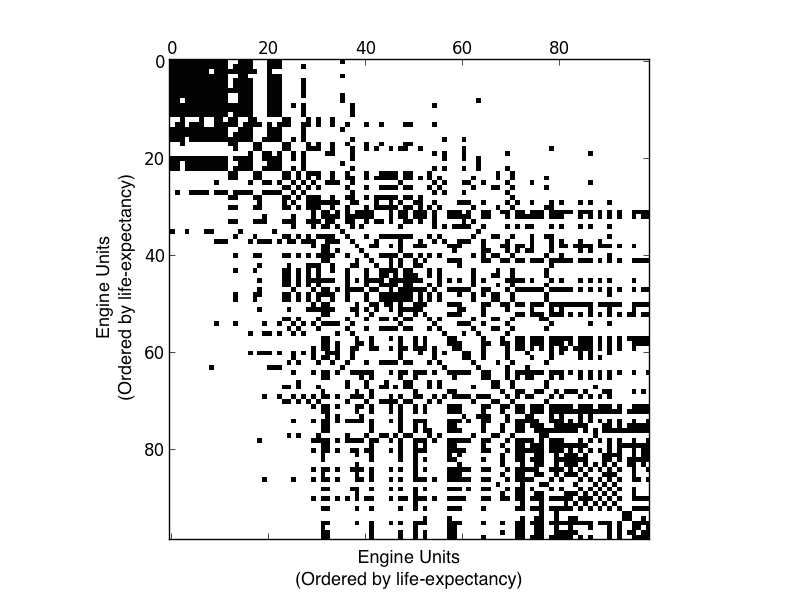}
  \caption{Pairwise cosine similarity between textual transforms for units in the dataset C-MAPSS1.Pairwise-similarities above the 75 percentile threshold have been shown in black. The textual transform was based only on the first 200 cycles for each unit}
  \vspace{-12pt}
  \label{fig:cluster}
\end{figure}

A key goal of prognostics is to enable efficient resource allocation for preventive maintenance.
After the initial 3 steps of encoding, discretization and tokenization to get the textual transform,
any well-known text clustering algorithm can be used on the transformed data containing
\emph{unit-as-a-document} to segment the fleet of units by their behavior. Figure~\ref{fig:cluster}
depicts pairwise cosine-similarity matrix of units transformed into documents. The appearance of block-matrices along the diagonal demonstrates that typically units with similar life-expectancies have similar textual profiles.


\section{Related Work}\label{sec:related-work}
The work in the area of prognostics and health management (PHM) can be divided into three
broad categories based on the time-horizon of predictions and prior knowledge of
faults or degradation. \emph{Anomaly detection} deals with detecting any unusual
changes in system chararteristics when the true cause of these changes may not be
known a priori or even understood. This approach is good and applicable for new
product lines whose degradation patterns may not be fully understood, and also to
monitor old and more studied systems to detect \emph{unknown} patterns and
events. These approaches typically employ unsupervised learning approaches to model
the nominal or healthy behavior and then flag systems deviating with respect to the
nominal. The output from an anomaly detection module can be used to trigger the
next stage of PHM for further investigation. The area of anomaly detection is
not limited to PHM and it has been applied in various domains, a summary of which 
appears in~\cite{Chandola2009}, and
~\cite{Pimentel2014}, while a review of anomaly detection for discrete sequences
can be found in~\cite{Chandola2012}

\emph{Fault detection \& Diagnostics} (FDD) involves prior knowledge of faults
under consideration known instances or physical understanding of its effect on
performance. The goal is to build models that can detect the presence of these
faults and categorize them. For equipment health management, when the effects of
a fault are well understood or when physically modeling the system
nominal behavior is feasible, physics-based approaches can be employed to help
detect and determine the root-cause of the fault. Supervised
data-driven approaches are used when the physics of failure is too complex to model or not
well understood. Good overall reviews of FDD can be found in~\cite{Venkat2003}.
The prediction time horizon for anomaly detection and fault detection \&
diagnostics is typically very short or current and they are used to determine
the current state of the system under consideration.

\emph{Prognostics} deals with forecasting equipment health state. This
problem is characteristically different from the previous two in terms of the
predicting time horizon and modeling of the temporal patterns of equipments
characteristics become imperative. Both model-based and data-driven approaches
have been used for this purpose. Model-based approaches such as particle
filtering~\cite{Marcos2005} make use of system-dynamics and fault propagation model to
estimate the health state of the system in the future. Data-driven approaches on
the other hand look to model the observed temporality due to system and fault
dynamics with minimal information. 
Review of prognostics methodologies
for rotating machines like an aircraft engines, batteries can be found in
found in~\cite{Lee2014, Saha2009}

Numerous discrete representations have been proposed for mining time series data. 
A detailed discussion on discretizing real-valued data appears in ~\cite{Garcia:2013:SDT:2478560.2478975}, 
and~\cite{Lin:2007:ESN:1285960.1285965} summarizes approaches specific to discretizing real-valued 
time-series data. 
Some of these approaches, such as Discrete Wavelet Transform (DWT)~\cite{1999:ETS:846218.847201}, Discrete Fourier Transform (DFT)~\cite{Faloutsos:1994:FSM:191839.191925}, Piecewise Linear/Constant models such as Piecewise Aggregate Approximation (PAA)~\cite{Yi:2000:FTS:645926.671689, Chakrabarti:2002:LAD:568518.568520} or APCA~\cite{Geurts:2001:PET:645805.670003} and Singular Value Decomposition (SVD)~\cite{Chakrabarti:2002:LAD:568518.568520}, offer real-valued representations that are not amenable to the textual transformation that we wish to achieve. The symbolic representation known as SAX (Symbolic Aggregate approXimation)~\cite{Lin:2003:SRT:882082.882086} is one approach suitable for our need of generating text documents. 
SAX and its variants are widely accepted as being suitable for common time-series mining tasks such as clustering~\cite{Ratanamahatana:2005:NBL:2140831.2140936}, classification~\cite{Siirtola:2011:ICA:2027478.2027690}, indexing~\cite{Shieh:2008:ISI:1401890.1401966}, summarization~\cite{Kumar:2005:PTV:1078022.1078137}, rare event detection~\cite{Keogh:2005:HSE:1106326.1106352}. 



\subsection{Key contributions} 
Although bag-of-words strategy for time-series has been explored before~\cite{Lin:2012:RST:2387443.2387446, saxvsm}, in these approaches, the time-series is first tokenized into fixed length subsequences and then converted into the corresponding SAX representation. A \emph{tf-idf} model is applied and 1-nearest neighbor classifier is used with cosine similarity as the distance metric. These approaches have been developed for univariate problems while ours is a multi-variate setting. Moreover, a distance metric based approach cannot filter irrelevant information. A regularized discriminative classifier such as a SVM or implicit feature selection using Random Forest is crucial for good performance. The deviance encoding step is also absent in these approaches, but it is very valuable for prognostics for encoding deterioration from normal. Also, tokenization used in these approaches is fixed length, while we offer an additional perspective of treating the time-series as a Markov chain for token length estimation. 

The focus of our approach is \emph{structural similarity} of time-series data, which is more relevant for discriminating dynamical patterns from engineering systems, as opposed to \emph{shape similarity} which has been primary focus of previous works such as SAX-based approaches. Also, by utilizing mixture of Gaussians in the deviance encoding step, we facilitate the discovery of multi-mode operation of industrial equipments. This is unlike previous works that assume unimodality. We present a supervised alternative (RMEP) for symbolization to leverage class labels to achieve more discriminative symbolization as opposed to unsupervised approaches like SAX.


Our approach of utilizing a symbolic
bag-of-words representation is novel for the prognostics application. Moreover,
the analogy of time-series as a journal of observations from an
industrial unit is a new perspective of formulating the prognostics
problem. This transformation enables leveraging significant advances
in scalability in text mining for prognostics and other
time-series analyses. Unlike most other approaches in prognostics that
are domain-specific, our methodology is data-driven allowing for a wider
application base.

Much of the recent research in scalability for big-data analysis has happened in the domain of text-mining and linear models that can be learnt via Stochastic Gradient Descent. Our transformation of the multivariate time-series into a textual form enables the field of prognostics to benefit from the recent advances in scalability and big-data analysis of textual data. 

\section{Conclusions and Future Work}\label{sec:conclusions}
We demonstrated a novel approach to transform multivariate time-series
data into a symbolic textual form that can directly leverage latest
research in text mining, for scalability, interpretability and
predictive performance. To this end, we have demonstrated successful
use cases that solve well known problems in prognostics using off-the
shelf text mining algorithms for classification. In addition to
accurately classifying industrial units based on their life
expectancy, the proposed approach can detect under-performance much
earlier. Moreover, the proposed approach is generic and
domain-agnostic allowing for a wider deployment.


Several directions are open for future exploration: (a) Integrated strategies
for alphabet size and token length selection, (b) framework to allow
incorporation of relevant domain knowledge if necessary, and (c)
effect of token order, e.g. n-grams, on predictive performance






\end{document}